\documentclass{midl} 

\usepackage{mwe} 
\usepackage{enumitem}
\usepackage[font=small,skip=0pt]{caption} 
\usepackage[rightcaption]{sidecap}
\usepackage{lscape}
\usepackage{rotating}
\usepackage{booktabs}
\jmlryear{2023}\jmlrworkshop{Full Paper -- MIDL 2023}\jmlrvolume{-- nnn}\editors{Accepted for publication at MIDL 2023}
\title[]{{ViT-AE++}: Improving Vision Transformer Autoencoder for Self-supervised Medical Image Representations}

\midlauthor{\Name{Chinmay Prabhakar\midljointauthortext{Contributed equally}\nametag{$^{1}$}} \Email{chinmay.prabhakar@uzh.ch} \\
\Name{Hongwei Bran Li\midlotherjointauthor\nametag{$^{1,2}$}} \Email{hongwei.li@tum.de}\\
\Name{Jiancheng Yang\nametag{$^{3}$}} \Email{jiancheng.yang@epfl.ch}\\
\Name{Suprosanna Shit\nametag{$^{2}$}} \Email{suprosanna.shit@tum.de}\\
\Name{Benedikt Wiestler  \midljointauthortext{Equal senior supervision}\nametag{$^{4}$}} \Email{b.wiestler@tum.de}\\
\Name{Bjoern H. Menze \nametag{$^{\dagger 1}$}} \Email{bjoern.menze@uzh.ch}\\
\addr $^{1}$ Department of Quantitative Biomedicine, University of Zurich, Switzerland \\
\addr $^{2}$ Department of Computer Science, Technical University of Munich, Germany \\
\addr $^{3}$ Computer Vision Laboratory, EPFL, Switzerland\\
\addr $^{4}$ Klinikum rechts der Isar, Technical University of Munich, Germany\\
}

\begin{document}

\maketitle

\begin{abstract}
Self-supervised learning has attracted increasing attention as it learns data-driven representation from data without annotations. Vision transformer-based autoencoder (\emph{ViT-AE}) \cite{he2021masked} is a recent self-supervised learning technique that employs a patch-masking strategy to learn a meaningful latent space. In this paper, we focus on improving \emph{ViT-AE} (nicknamed \emph{ViT-AE$+$$+$}) for a more effective representation of both 2D and 3D medical images. We propose two new loss functions to enhance the representation during the training stage. The first loss term aims to improve self-reconstruction by considering the structured dependencies and hence \emph{indirectly} improving the representation. The second loss term leverages contrastive loss to \emph{directly} optimize the representation from two randomly masked views.
As an independent contribution, we extended \emph{ViT-AE$+$$+$} to a 3D fashion for volumetric medical images. 
We extensively evaluate \emph{ViT-AE$+$$+$} on both natural images and medical images, demonstrating consistent improvement over vanilla \emph{ViT-AE} and its superiority over other contrastive learning approaches. Our code is available at \url{https://github.com/chinmay5/vit_ae_plus_plus.git}


\end{abstract}

\begin{keywords}
representation; self-supervised learning; masked vision transformer
\end{keywords}

\section{Introduction}


Self-supervised representation learning (SSRL), especially recent contrastive learning-based methods \cite{chen2020simple,he2020momentum,oord2018representation}, is a promising technique that can learn informative data representations from data without labels. It is particularly useful and relevant in medical imaging, where labeled data are often scarce for traditional supervised training and the high cost of manual labeling that relies on domain knowledge. 

Contrastive learning-based approaches aim to attract \emph{similar} image pairs and rebuff \emph{dissimilar} image pairs. A similar image pair can be generated by two augmented views of one image, and the other image samples can be used to construct dissimilar image pairs. In medical imaging, \emph{SSRL} are mainly used for two purposes: 1) pre-training deep networks for transfer learning or for network initialization \cite{zhou2019models,chaitanya2020contrastive,zeng2021positional}, 2) extracting meaningful information from data \cite{li2021imbalance,dufumier2021contrastive} which is the downstream task for applying \emph{SSRL} in this work. 

As opposed to contrastive learning, the recent vision transformer autoencoder (\emph{ViT-AE}) approach \cite{he2021masked} is different from the above methods in principle. 
It randomly masks the sequential image patches and learns to reconstruct the original image using an autoencoder and vision transformer-based architecture. 
They demonstrate that such a simple reconstruction can facilitate the vision transformer to learn effective image representations in a self-supervised fashion.
Although \emph{ViT-AE} achieves promising results in the natural image domain, we argue that two components of this framework could be further optimized. 

First, in the vanilla \emph{ViT-AE} pipeline, it computes pixel-wise reconstruction loss of the autoencoder and does not consider structured dependencies of the reconstruction, which might limit to capture of semantic features. 
For example, medical images contain rich texture and morphology, such structural information across pixels is important to complement traditional pixel-wise reconstruction (often with a $\mathcal{L}$-2 norm loss).

Second, since the representation is \emph{indirectly} learned by a self-reconstruction via an autoencoder, there might be room to optimize the target representation directly. For example, contrastive learning-based methods \cite{chen2021exploring,chen2020simple} straightforwardly match the target representation from two augmented views. 
\noindent
\paragraph{Contributions.} (1) We introduce an auxiliary reconstruction task that considers structural dependencies to complement the pixel-wise reconstruction. (2) We unite two paradigms of contrastive learning-based and autoencoder-based methods and enjoy the benefits of both. (3) In extensive experiments, we demonstrate that both 2D and 3D \emph{ViT-AE$+$$+$} outperform the vanilla \emph{ViT-AE} and its superiority over other contrastive learning approaches, setting up a strong baseline for learning self-supervised medical image representation.

\section{Related Work}
We mainly discuss prior works that are related to the technical contributions of \emph{ViT-AE$+$$+$}. 
\paragraph{Auxiliary loss functions.} In image segmentation tasks, auxiliary losses can regularize network training by serving as `deep supervision' \cite{lee2015deeply,zhao2017pyramid}. 
\citet{liu2021generic} introduce a generic perceptual loss for dense prediction tasks. 
They argue that leveraging an auxiliary task that considers structural dependency can benefit various dense prediction tasks. Inspired by this work, we introduce an auxiliary reconstruction task to the autoencoder. In addition to perceptual loss, we develop a new edge-aware loss considering the richness of texture in medical images. 

\paragraph{Contrastive loss.} Contrastive loss, such as \emph{SimCLR} \cite{chen2020simple} directly optimizes the image representation by maximizing the agreement between two augmented views. The asymmetric networks such as \emph{SimSiam} \cite{chen2020exploring} and \emph{BYOL} \cite{grill2020bootstrap} follow a similar idea but only make use of positive pairs with two shared encoders. Considering the training efficiency, we borrow the loss design from \emph{SimSiam} \cite{chen2020exploring}. Differently, we not only use random augmentation but also random masking to obtain hard training samples, analogous to random cropping in \emph{CNN}-based backbone.

\section{Method}
\begin{figure}[t!]
    \centering
    \includegraphics[width=0.95\textwidth]{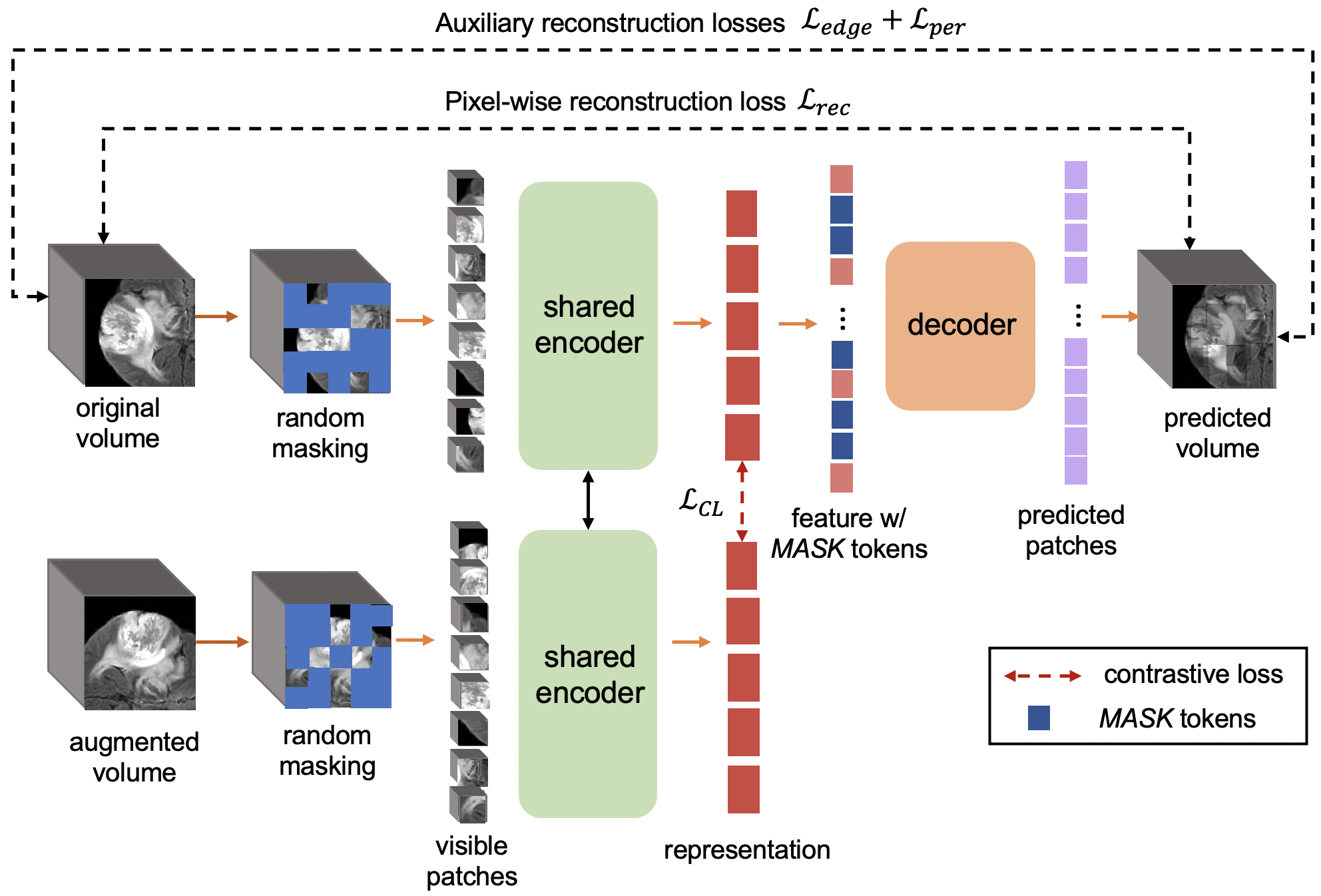}

    \caption{\small Our proposed \emph{ViT-AE++} framework. The upper half shows the training procedure of the autoencoder. An MRI volume is parsed and masked randomly and the visible patches are passed through to the encoder. Placeholder tokens for the masked patches are introduced \textit{after} the encoder (referred to as the \textit{MASK} tokens). The full set of encoded features and these placeholder tokens are passed to a decoder which reconstructs the whole MRI volume. We use a pixel-wise reconstruction loss and an auxiliary loss that concerns perceptual features (Eq.~\ref{eq:perc_loss}) and edge map (Eq.~\ref{eq:sobel_op}). The lower half demonstrates the optimization with a contrastive loss which is based on two randomly augmented and masked views. The representations from the two views are matched with cosine similarity loss (Eq.~\ref{negative_cos}).}
    \label{fig:main_ViT}
    \vspace{-0.3cm}
\end{figure}

\paragraph{Overview.} \label{method}
The objective is to learn a good domain-specific representation of 3D volumes using an autoencoder without labels. Consider an autoencoder with encoder ${E}$, decoder ${D}$, and a function $m(\cdot)$ for masking input patches in a vision transformer architecture. Given an image volume $X$, it is decomposed into $k$ sequential patches with a size of $n$$\times$n$\times$$n$.
Then a random mask is applied on the sequential input patches to mask away $p$\% of the patches. The remaining visible patches are referred to as $X^{*}$. 
The visible patches $X^{*}$ are fed to the encoder ${E}$ to extract features. 
For each missing patch, a token with 3D positional encoding is assigned to indicate the presence of such patches (called \textit{MASK} tokens).
The features of the visible patches along with  \textit{MASK} tokens are passed to the decoder. 
The decoder ${G}$ predicts image intensities for these \textit{MASK} tokens and thereby reconstructs the whole volume, i.e.,
$X\approx {D}({E}(X^{*}))$. We introduce an auxiliary reconstruction task with a compound loss function $\mathcal{L}_{per}+\mathcal{L}_{edge}$ as shown in Fig.~\ref{fig:main_ViT}. The new loss is designed to capture high-order properties to complement the pixel-wise loss. To further enhance the target representation, we adopt contrastive loss to maximize the agreement from two random masked views. Fig.~\ref{fig:main_ViT} shows the schematic view of our architecture. In the following sections, we explain each component in detail. 


\noindent
\paragraph{Vanilla \emph{ViT-AE} with pixel-wise loss.} \emph{ViT-AE} takes partial observations and reconstructs the original input. 
A random masking strategy is used to mask $p$\% of the input volume. Notably, $p$ is a hyperparameter which will be discussed in a later section. The visible patches $X^{*}$ are fed through encoder ${E}$. The decoder ${D}$ reconstructs $X$ from ${E} (X^{*})$
using a mean-squared-error loss: $\mathcal{L}_{rec} = ||{X} - {D}({E} (X^{*})||_2$.
\paragraph{Auxiliary compound loss.} Original \emph{ViT-AE} uses the pixel-wise loss for training as mentioned above. 
In the medical image domain, perceptual features and edges encode meaningful semantics signals \cite{shen2017boundary}. We employ a compound loss with deep high-level features and low-level edge-based features to enforce the network to use this supervision signal during training. 

To exact deep multi-level features, we introduce \emph{VGG}-based perceptual loss~\cite{johnson2016perceptual} to compute the feature similarity at multiple levels over $n$ 2D slices of one volume. 

\begin{equation} \label{eq:perc_loss}
	\mathcal{L}_{per} = \sum_{i=0}^{n}||VGG(x_{i}) - VGG(\hat{x_{i}})||_{2}
	\end{equation}
where $VGG(\cdot)$ denotes multi-level features from the pre-trained \emph{VGG} network using the same layers in \citep{johnson2016perceptual}. $x_{i}$ and $\hat{x_{i}}$ denote the input image and the reconstructed image, respectively. 

For low-level features, a 3D \emph{Sobel} filter function \cite{kanopoulos1988design} $Sobel(\cdot)$ is used to compute the gradients from three directions. 
Firstly, the input volume is convolved with a fixed \emph{Sobel} filter to generate gradients in the axial, coronal, and sagittal directions. The norm of these gradients is the edge map representation (Eq.~\ref{eq:sobel_op}). Please see the details in the Appendix.

The edge loss between original volume $X$ and the reconstruction $\hat{X}$ is formulated as: 
\begin{equation} \label{eq:sobel_op}
\mathcal{L}_{edge} = ||\emph{Sobel}(X) -  \emph{Sobel}(\hat{X})||_2
\end{equation}




\paragraph{Contrastive loss.} Along with the self-reconstruction, a contrastive loss is further introduced to enhance the target representation. It tries to match the feature representations between the two views of visible patches, denoted as $f_{1}$ and $f_{2}$. We used negative cosine distance as the loss function and the `stopping gradient' optimization in \cite{chen2020exploring}. 

\begin{equation}
\label{negative_cos}
\mathcal{L}_{CL}=-\frac{f_{1}}{\left\|f_{1}\right\|_{2}} \cdot \frac{{f_{2}}}{\left\|f_{2}\right\|_{2}},
\end{equation}
\paragraph{\emph{ViT-AE$+$$+$}.} The final optimization objective of \emph{ViT-AE$+$$+$} is
\begin{equation} \label{eq:final_loss}
L = \mathcal{L}_{rec} + \lambda_1 \mathcal{L}_{per} + \lambda_2 \mathcal{L}_{edge} + \mathcal{L}_{CL}.
\end{equation}

where $\lambda_1$ and $\lambda_2$ are hyperparameters and initially set to 0.01 and 10 respectively considering the loss scale. $\lambda_2$ is decreased linearly to stabilize the training process. The motivation for the linear scaling and other hyperparameters are discussed in the Results section.

\section{Experiment Setup}
We evaluate \emph{ViT-AE$+$$+$} extensively on public natural and medical image datasets, focusing on 3D brain MRI datasets.
\paragraph{2D Datasets.} \emph{CIFAR-10}, \emph{CIFAR-100} and Tiny Imagenet-100 \cite{russakovsky2015imagenet} are popular natural datasets to benchmark representation learning methods. In addition, we evaluate our method on a 2D chest X-ray dataset \cite{kermany2018identifying}. It consists of 5118 images for training, 119 for validation, and 626 for testing. The task is to identify pneumonia based on chest X-ray images. All images are resized to 224$\times$224 pixels.

\paragraph{3D Datasets.} 
To highlight the effectiveness of 3D \emph{ViT-AE$+$$+$}, we performed experiments on two public 3D datasets: 1) a multi-center MRI dataset (\emph{BraTS}) \cite{menze2014multimodal,bakas2018identifying} including 326 patients with a brain tumor. 2) The Erasmus Glioma Database (\emph{EGD}) \cite{van2021erasmus} which consists of 768 MRI scans. Both datasets include FLAIR, T1, T2, and T1-c with a uniform voxel size 1$\times$1$\times$1 \emph{mm$^3$}. We train individual models and evaluate on the two datasets separately. 
The effectiveness of the learned representations is evaluated on two down-stream classification tasks: a) For \emph{BraTS}, discriminating high-grade and low-grade tumor, and b) for \emph{EGD}, predicting isocitrate dehydrogenase (IDH) mutation status (0 or 1). 

We use the segmentation masks for both datasets to get its centroid and generate a 3D bounding box of 96$\times$96$\times$96 to localize the tumor. If the bounding box exceeds the original volume, the out-of-box region is padded with background intensity. Four modalities are concatenated to serve as a multi-modal input. 
The intensity range of all image volumes was rescaled to [0,~255] to guarantee the success of intensity-based augmentations. 

\noindent
\paragraph{Architecture and training configuration.} In this sub-section, we explain each component in the architecture and the training settings in detail.
{\it \flushleft Encoder and decoder.} 
Working on only the visible patches allows the encoder to process a fraction of 3D volumes with less GPU memory. We extend a standard ViT~\cite{dosovitskiy2021image} to be 3D architecture as the encoder. 3D patches are embedded using a linear projection with added positional encoding with 3D coordinates. 12 encoder blocks, each with an embedding dimension of 768, are used. The decoder reconstructs the volume using features from the encoder and the \textit{MASK} tokens mentioned in Sec. \ref{method}. Each \textit{MASK} token is a shared learned vector that indicates the presence of a missing patch to be predicted. We add 3D positional encoding to all the tokens for sequential prediction \cite{devlin2019bert}. The decoder is realized using 8 transformer blocks, each with an embedding dimension of 512. After training, we only use the encoder as a feature extractor. The 3D position encoding is shown in the Appendix. 
{\it \flushleft End-to-end training of \emph{ViT-AE$+$$+$} with contrastive loss.} \emph{ViT-AE$+$$+$} is trained for 1000 epochs using~\emph{AdamW} optimizer~\cite{loshchilov2019decoupled} with 0.05 weight decay. The base learning rate is 1e-3. The learning rate is annealed using cosine decay~\cite{loshchilov2017sgdr}. The batch size is set to 4, adjusted to maximize GPU memory with an Nvidia RTX 3090. We use 40 epochs for our warm-up schedule \cite{goyal2018accurate}.
\textit{Gamma correction, affine transform and Gaussian noise} augmentations are used to generate the second "view" of the input volume. The two views are randomly masked and passed through the shared encoder ${E}$. Features generated by the encoder are compared using their cosine similarity (Eq.~\ref{negative_cos}).

\noindent
\paragraph{Evaluation strategy, classifier, and metrics.} 
For the \emph{EGD} dataset, since there are 307 unlabeled images, we pre-train on the unlabeled data and perform the downstream task on the labeled dataset. For the other 2D and 3D datasets, the training of the proxy task and the target downstream task use the same dataset. 

For evaluation on \emph{BraTS} and \emph{EGD} datasets, we follow the standard strategy to evaluate the quality of the pre-trained representations by training a \emph{supervised} linear support vector machine classifier on the training set and then evaluating it on the test set. We use the sensitivity, specificity and Area Under the Receiver Operating Characteristic Curve (AUC) as the evaluation metrics.
We use \emph{stratified five-fold nested cross-validation} to reduce selection bias and validate each model. In each fold, we randomly sample 80\% subjects from each class as the training set and the remaining 20\% for each class as the test set. Furthermore, 20\% of training data is separated and used exclusively to optimize the hyper-parameters within each fold. 

\label{par:lin_prob}For evaluation on \emph{CIFAR-10}, \emph{CIFAR-100}, Tiny Imagenet-100, and chest X-ray datasets, we use the linear probing strategy \cite{he2021masked}. The decoder module is discarded and encoder weights are frozen. Thus, the encoder acts as a feature extractor. A supervised linear classifier is trained on the frozen representations. We use the pre-defined train and test splits. 20 \% of the training data is separated for hyper-parameter tuning. We report the classification accuracy for the test split for all the datasets.

\section{Results}

\noindent

\paragraph{ViT-AE$++$ $vs.$ ViT-AE on 2D datasets.}
To quantify the effectiveness of our proposed loss functions, we compare our method against the vanilla ViT-AE on \emph{CIFAR-10}, \emph{CIFAR-100}, Tiny Imagenet-100 and \emph{chest X-ray} datasets. We use the linear probing evaluation protocol and report the classification accuracy on the test sets. Please note that the VGG-perceptual loss for three natural image datasets is not a pre-trained VGG but one with randomized weights. This is because we do not wish any supervised signals involved during the self-supervised training stage. Tab. \ref{tab:results_nat_images} shows that the features learned using ViT-AE$++$ consistently perform better than their vanilla counterpart. It indicates that ViT-AE$++$ can serve as a strong baseline for both natural and medical image domains. 

\begin{table*}[htpb]
  \caption[table: nat_img_comparison]{\small Comparison of ViT-AE$++$ and ViT-AE on four 2D datasets. We can observe consistent improvements in classification accuracy, benefiting from the new loss functions.}
  \label{tab:results_nat_images}
  \centering
  \setlength{\tabcolsep}{2mm}{
   \begin{tabular}{c | c| c| c | c}
    \toprule 

     Method & CIFAR-10 & CIFAR-100 & Tiny ImageNet-100 & chest X-ray\\
     
    \hline
     ViT-AE & 94.10 & 75.61 & 70.42 & 95.20 \\
     ViT-AE++ (ours) & \textbf{95.40}  & \textbf{78.82} & \textbf{72.09} & \textbf{95.60}\\
     \bottomrule
  \end{tabular}
}
\end{table*}

\paragraph{ViT-AE$++$ $vs.$ other methods on 3D datasets.}
To focus on 3D medical image representation on small-scale datasets, we further evaluate the 3D version of ViT-AE$++$ and compare it with other state-of-the-art self-supervised methods (especially contrastive learning-based ones) on two classification tasks: a) discrimination of low-grade and high-grade brain tumors and b) prediction of IDH mutation status. 

We observe that {ViT-AE$++$} behaves differently on two 3D datasets as shown in Table~\ref{tab:results_all_ssl}. On \emph{BraTS}, ViT-AE$++$ is competitive to existing contrastive learning-based methods including MoCO-v3 \cite{chen2021empirical} and SimSiam \cite{chen2020simple}, achieving AUCs of 0.767 \emph{vs.} 0.795 and 0.767 \emph{vs.} 0.771 respectively. This may be caused by overfitting since \emph{BraTS} has only 260 samples in each training fold. On \emph{EGD} which contains two times amount of training samples, ViT-AE$++$ outperforms the two methods by a large margin, achieving AUCs of 0.846 \emph{vs.} 0.734 and 0.846 \emph{vs.} 0.741 respectively. Notably, the proposed new loss functions consistently improve the representation in the ViT-AE framework. As shown in Table~\ref{tab:ablation}, we observe that the auxiliary compound loss and contrastive loss improve vanilla ViT-AE significantly. Especially, to demonstrate the effectiveness of our proposed auxiliary compound loss, we visualize the reconstruction results by different combinations of reconstruction loss functions shown in Figure \ref{fig:reconstruction}. 







\begin{table}[!htb]
\scriptsize
    \begin{minipage}{.5\linewidth}
      \caption[table: Comparison_all_methods]{\small Comparison of ViT-AE$++$ and other \\ self-supervised methods.}
        \label{tab:results_all_ssl}
      \centering
        \begin{tabular}{l |  c |  c }
        \toprule
         &{\textit{BraTS}} & {\textit{EGD}} \\
        {Methods}
         & \textit{AUC}  & \textit{AUC} \\
        \hline
         MoCO-v3  & \textbf{0.795} $\pm$ 0.065 & 0.734 $\pm$ 0.048 \\
         SimSiam & 0.771 $\pm$ 0.065 & 0.741 $\pm$ 0.039\\
         Vallina ViT-AE & 0.696 $\pm$ 0.079 & 0.828 $\pm$ 0.036\\ 
        \hline
         ViT-AE++(ours) & {0.767} $\pm$ 0.068&\textbf{0.846} $\pm$ 0.034\\
    
        \bottomrule
      \end{tabular}
    \end{minipage}%
    \begin{minipage}{.5\linewidth}
      \centering
        \caption[table: Comparison]{\small Ablation study of the auxiliary loss and contrastive loss on the BraTS dataset.}
        \label{tab:ablation}
        \begin{tabular}{l | c }
    \toprule
 
     &\multicolumn{1}{c}{\textit{BraTS}} \\
    {Methods}
    ~ & \textit{AUC}       \\

\hline
     ViT$+\mathcal{L}_{CL}$ & 0.680 $\pm$ 0.091 \\
     ViT-AE  & 0.696 $\pm$ 0.057\\ 
     ViT-AE+$\mathcal{L}_{edge}$  & 0.704 $\pm$ 0.065\\ 
     ViT-AE+$\mathcal{L}_{Per}$ & 0.721 $\pm$ 0.085\\
     ViT-AE+$\mathcal{L}_{edge}$+$\mathcal{L}_{Per}$ & 0.734 $\pm$ 0.088\\
     ViT-AE+$\mathcal{L}_{edge}$+$\mathcal{L}_{Per}$+$\mathcal{L}_{CL}$  & \textbf{0.767} $\pm$ 0.069\\
     \bottomrule
  \end{tabular}
    \end{minipage} 
\end{table}

\paragraph{Does backbone matter?}
Our proposed framework used the vision transformer (ViT) as our feature extraction backbone. One might argue that the observed performance improvement is a consequence of using the ViT and not the proposed auxiliary training objectives. In such a scenario, plugging in the ViT in other self-supervised representation learning methods such as MoCOv3 \cite{he2020momentum} and SimSiam \cite{chen2021exploring} should lead to a corresponding increase in representation strength. Tab. \ref{tab:results_all_ssl_vit} shows the results on the \emph{BraTs} and \emph{EGD} datasets. The self-supervised representation features learned using vision transformers perform poorly compared to their ResNet counterpart. We believe this results from overfitting due to a large number of model parameters in ViT. Thus, we conclude that a simple replacement of the feature extractor \emph{does not} guarantee superior performance.

\begin{table*}[!htpb]
  \caption[table: Comparison_all_methods]{\small Performances of vision transforms trained using different self-supervised representation learning methods.}
  \label{tab:results_all_ssl_vit}
  \centering
  \setlength{\tabcolsep}{2mm}{
   \begin{tabular}{l | c | c }
    \toprule

    &\multicolumn{1}{c|}{\textit{BraTS}} & \multicolumn{1}{c}{\textit{EGD}} \\
    {Methods}
     ~ &  \textit{AUC} &  \textit{AUC}      \\
     \hline
     MoCOv3 & 0.795 $\pm$ 0.057 &  0.734 $\pm$ 0.048 \\
     SimSiam  & 0.771 $\pm$ 0.065 & 0.741 $\pm$ 0.039\\
    \hline
     SimSiam (ViT)  & 0.605 $\pm$ 0.088 & 0.774 $\pm$ red{0.033}\\
     MoCov3 (ViT) & 0.714 $\pm$ 0.069 &  0.798 $\pm$ 0.029\\
 

\bottomrule
  \end{tabular}
}
\end{table*}

\paragraph{Analysis of hyperparameters.}

\begin{figure}[t!]
    \centering
    \includegraphics[width=0.95\textwidth]{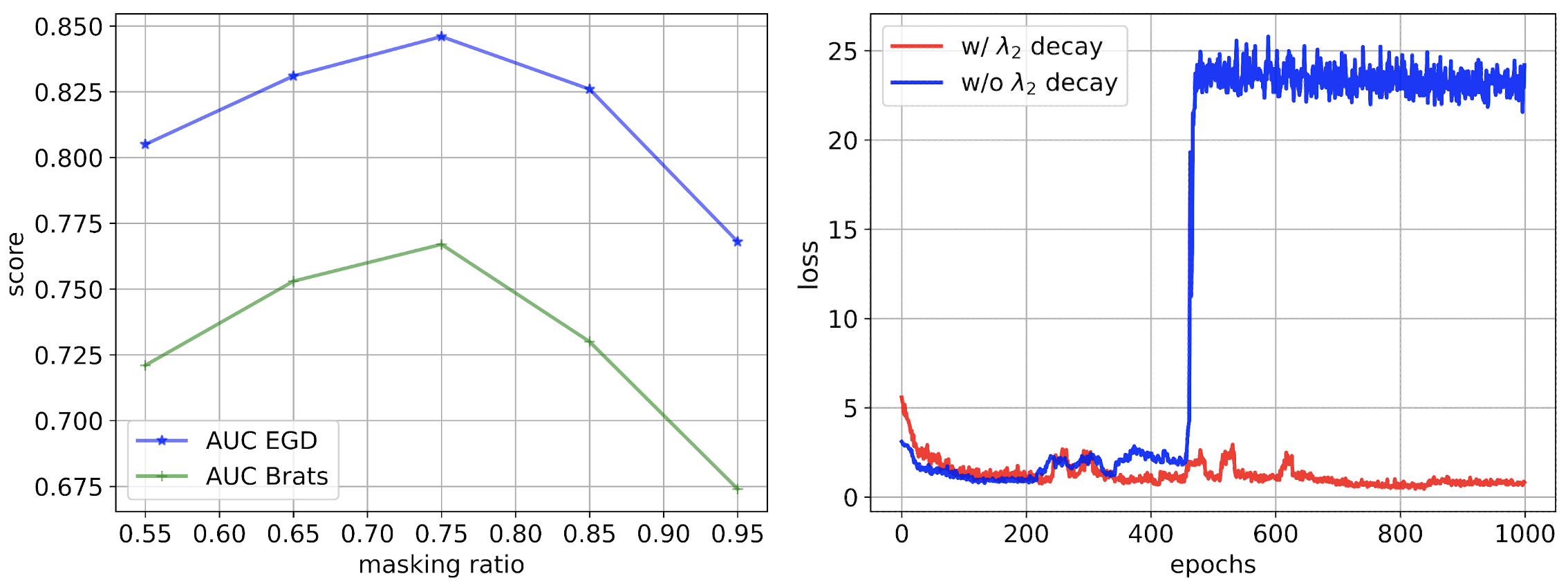}
    \caption{\small Effects of two key parameters. \textbf{Left:} Masking ratio \emph{vs.} AUC on the \textit{BraTS} dataset. The AUC and specificity increase with higher masking ratio, peaking at 0.75, after which they begin to decrease. \textbf{Right:} Training loss with Eq. \ref{eq:final_loss} over epochs with and without weight decay of $\lambda_2$ for $\mathcal{L}_{edge}$. Without decay (blue), the loss may explode and the training could not converge; With decay (red), training is stable and converges.}
    \label{fig:ablation}
\end{figure} 

We analyze two critical parameters in the proposed framework which affect the representation and training stability.
{\it \flushleft Effect of masking ratio $p$.} 
The masking ratio $p$ 
determines what percentage of the input volume is masked away. This, in turn, controls the difficulty of the reconstruction task. A higher value of $p$ implies fewer visible patches $X^{*}$, which makes reconstruction more challenging. On the other hand, a low $p$ can allow the model to extrapolate between visible patches, thus not learning good features. To find a suitable value of $p$, we run experiments with $p \in [0.55, 0.65, 0.75, 0.85, 0.95]$. The results for the \textit{BraTS} dataset are summarized in Figure \ref{fig:ablation}. We observe that the highest AUC value is obtained at $p=0.75$. 
In our experiments, the optimal value was $p=0.75$ for both datasets.
{\it \flushleft Weight decay for edge-based loss.} The edge-based loss weight $\lambda_2$ in Eq.~\ref{eq:final_loss} was decreased linearly with number of epochs. We observed that gradually decreasing $\lambda_2$ was essential for the training stability. 
One possible explanation for the instability is related to the one-to-multiple mapping of the reconstruction task. While the edge map $\mathcal{L}_{edge}$ provides strong additional supervision signals for reconstruction, it is sensitive to noise artifacts which can cause large losses. Fitting to such artifacts destabilizes the training process and hampers the embedding features. To avoid this overfitting, we start the training with $\lambda_2 = 0.01$ and linearly decrease it with each epoch. This allows the network to learn structural information initially and gradually focus more on perceptual similarity. The loss plot is shown in Fig.~ \ref{fig:ablation}. Without a weight decay of $\lambda_2$, the loss tends to diverge. When using a linear decay, the training is stable and gradually converges.
\section{Discussion and conclusion}
We proposed ViT-AE$++$ to improve the existing vision transformer-based self-supervised learning approach, which is orthogonal to the existing contrastive learning-based approaches. We introduce a new auxiliary reconstruction loss to the vision transformer autoencoder and extend it with a contrastive loss. We demonstrate that our proposed method is superior to vanilla ViT-AE and competitive to contrastive learning-based methods
We hope our work can provide a new perspective for representation learning in medical imaging and advance self-supervised features to the next level. In this work, we focus on learning representation on a single dataset and evaluate it in a downstream task using the same dataset. In future work, we will investigate learning generalizable self-supervised representations immune to common domain shifts (e.g., caused by image acquisition).

\midlacknowledgments{We would like to thank the Helmut Horten Foundation for supprting our research. Additionally, Hongwei Bran Li is supported by an Nvidia Academic GPU grant and Forschungskredit (grant No. K-74851-01-01) from the University of Zurich.}

\bibliography{midl23_023}

\begin{thebibliography}{30}
\providecommand{\natexlab}[1]{#1}
\providecommand{\url}[1]{\texttt{#1}}
\expandafter\ifx\csname urlstyle\endcsname\relax
  \providecommand{\doi}[1]{doi: #1}\else
  \providecommand{\doi}{doi: \begingroup \urlstyle{rm}\Url}\fi

\bibitem[Bakas et~al.(2018)Bakas, Reyes, Jakab, Bauer, Rempfler, Crimi,
  Shinohara, Berger, Ha, Rozycki, et~al.]{bakas2018identifying}
Spyridon Bakas, Mauricio Reyes, Andras Jakab, Stefan Bauer, Markus Rempfler,
  Alessandro Crimi, Russell~Takeshi Shinohara, Christoph Berger, Sung~Min Ha,
  Martin Rozycki, et~al.
\newblock Identifying the best machine learning algorithms for brain tumor
  segmentation, progression assessment, and overall survival prediction in the
  brats challenge.
\newblock \emph{arXiv preprint arXiv:1811.02629}, 2018.

\bibitem[Chaitanya et~al.(2020)Chaitanya, Erdil, Karani, and
  Konukoglu]{chaitanya2020contrastive}
Krishna Chaitanya, Ertunc Erdil, Neerav Karani, and Ender Konukoglu.
\newblock Contrastive learning of global and local features for medical image
  segmentation with limited annotations.
\newblock \emph{arXiv preprint arXiv:2006.10511}, 2020.

\bibitem[Chen et~al.(2020)Chen, Kornblith, Norouzi, and Hinton]{chen2020simple}
Ting Chen, Simon Kornblith, Mohammad Norouzi, and Geoffrey Hinton.
\newblock A simple framework for contrastive learning of visual
  representations.
\newblock \emph{arXiv preprint arXiv:2002.05709}, 2020.

\bibitem[Chen and He(2020)]{chen2020exploring}
Xinlei Chen and Kaiming He.
\newblock Exploring simple siamese representation learning.
\newblock \emph{arXiv preprint arXiv:2011.10566}, 2020.

\bibitem[Chen and He(2021)]{chen2021exploring}
Xinlei Chen and Kaiming He.
\newblock Exploring simple siamese representation learning.
\newblock In \emph{Proceedings of the IEEE/CVF Conference on Computer Vision
  and Pattern Recognition}, pages 15750--15758, 2021.

\bibitem[Chen et~al.(2021)Chen, Xie, and He]{chen2021empirical}
Xinlei Chen, Saining Xie, and Kaiming He.
\newblock An empirical study of training self-supervised vision transformers.
\newblock In \emph{Proceedings of the IEEE/CVF International Conference on
  Computer Vision}, pages 9640--9649, 2021.

\bibitem[Devlin et~al.(2019)Devlin, Chang, Lee, and Toutanova]{devlin2019bert}
Jacob Devlin, Ming-Wei Chang, Kenton Lee, and Kristina Toutanova.
\newblock Bert: Pre-training of deep bidirectional transformers for language
  understanding, 2019.

\bibitem[Dosovitskiy et~al.(2021)Dosovitskiy, Beyer, Kolesnikov, Weissenborn,
  Zhai, Unterthiner, Dehghani, Minderer, Heigold, Gelly, Uszkoreit, and
  Houlsby]{dosovitskiy2021image}
Alexey Dosovitskiy, Lucas Beyer, Alexander Kolesnikov, Dirk Weissenborn,
  Xiaohua Zhai, Thomas Unterthiner, Mostafa Dehghani, Matthias Minderer, Georg
  Heigold, Sylvain Gelly, Jakob Uszkoreit, and Neil Houlsby.
\newblock An image is worth 16x16 words: Transformers for image recognition at
  scale, 2021.

\bibitem[Dufumier et~al.(2021)Dufumier, Gori, Victor, Grigis, Wessa, Brambilla,
  Favre, Polosan, Mcdonald, Piguet, et~al.]{dufumier2021contrastive}
Benoit Dufumier, Pietro Gori, Julie Victor, Antoine Grigis, Michele Wessa,
  Paolo Brambilla, Pauline Favre, Mircea Polosan, Colm Mcdonald, Camille~Marie
  Piguet, et~al.
\newblock Contrastive learning with continuous proxy meta-data for 3d mri
  classification.
\newblock In \emph{International Conference on Medical Image Computing and
  Computer-Assisted Intervention}, pages 58--68. Springer, 2021.

\bibitem[Goyal et~al.(2018)Goyal, Dollár, Girshick, Noordhuis, Wesolowski,
  Kyrola, Tulloch, Jia, and He]{goyal2018accurate}
Priya Goyal, Piotr Dollár, Ross Girshick, Pieter Noordhuis, Lukasz Wesolowski,
  Aapo Kyrola, Andrew Tulloch, Yangqing Jia, and Kaiming He.
\newblock Accurate, large minibatch sgd: Training imagenet in 1 hour, 2018.

\bibitem[Grill et~al.(2020)Grill, Strub, Altch{\'e}, Tallec, Richemond,
  Buchatskaya, Doersch, Avila~Pires, Guo, Gheshlaghi~Azar,
  et~al.]{grill2020bootstrap}
Jean-Bastien Grill, Florian Strub, Florent Altch{\'e}, Corentin Tallec, Pierre
  Richemond, Elena Buchatskaya, Carl Doersch, Bernardo Avila~Pires, Zhaohan
  Guo, Mohammad Gheshlaghi~Azar, et~al.
\newblock Bootstrap your own latent-a new approach to self-supervised learning.
\newblock \emph{Advances in Neural Information Processing Systems}, 33, 2020.

\bibitem[He et~al.(2020)He, Fan, Wu, Xie, and Girshick]{he2020momentum}
Kaiming He, Haoqi Fan, Yuxin Wu, Saining Xie, and Ross Girshick.
\newblock Momentum contrast for unsupervised visual representation learning.
\newblock In \emph{Computer Vision and Pattern Recognition}, pages 9729--9738,
  2020.

\bibitem[He et~al.(2021)He, Chen, Xie, Li, Doll{\'a}r, and
  Girshick]{he2021masked}
Kaiming He, Xinlei Chen, Saining Xie, Yanghao Li, Piotr Doll{\'a}r, and Ross
  Girshick.
\newblock Masked autoencoders are scalable vision learners.
\newblock \emph{arXiv preprint arXiv:2111.06377}, 2021.

\bibitem[Johnson et~al.(2016)Johnson, Alahi, and
  Fei-Fei]{johnson2016perceptual}
Justin Johnson, Alexandre Alahi, and Li~Fei-Fei.
\newblock Perceptual losses for real-time style transfer and super-resolution.
\newblock In \emph{European conference on computer vision}, pages 694--711.
  Springer, 2016.

\bibitem[Kanopoulos et~al.(1988)Kanopoulos, Vasanthavada, and
  Baker]{kanopoulos1988design}
Nick Kanopoulos, Nagesh Vasanthavada, and Robert~L Baker.
\newblock Design of an image edge detection filter using the sobel operator.
\newblock \emph{IEEE Journal of solid-state circuits}, 23\penalty0
  (2):\penalty0 358--367, 1988.

\bibitem[Kermany et~al.(2018)Kermany, Goldbaum, Cai, Valentim, Liang, Baxter,
  McKeown, Yang, Wu, Yan, et~al.]{kermany2018identifying}
Daniel~S Kermany, Michael Goldbaum, Wenjia Cai, Carolina~CS Valentim, Huiying
  Liang, Sally~L Baxter, Alex McKeown, Ge~Yang, Xiaokang Wu, Fangbing Yan,
  et~al.
\newblock Identifying medical diagnoses and treatable diseases by image-based
  deep learning.
\newblock \emph{cell}, 172\penalty0 (5):\penalty0 1122--1131, 2018.

\bibitem[Lee et~al.(2015)Lee, Xie, Gallagher, Zhang, and Tu]{lee2015deeply}
Chen-Yu Lee, Saining Xie, Patrick Gallagher, Zhengyou Zhang, and Zhuowen Tu.
\newblock Deeply-supervised nets.
\newblock In \emph{Artificial intelligence and statistics}, pages 562--570.
  PMLR, 2015.

\bibitem[Li et~al.(2021)Li, Xue, Chaitanya, Luo, Ezhov, Wiestler, Zhang, and
  Menze]{li2021imbalance}
Hongwei Li, Fei-Fei Xue, Krishna Chaitanya, Shengda Luo, Ivan Ezhov, Benedikt
  Wiestler, Jianguo Zhang, and Bjoern Menze.
\newblock Imbalance-aware self-supervised learning for 3d radiomic
  representations.
\newblock In \emph{International Conference on Medical Image Computing and
  Computer-Assisted Intervention}, pages 36--46. Springer, 2021.

\bibitem[Liu et~al.(2021)Liu, Chen, Chen, Yin, and Shen]{liu2021generic}
Yifan Liu, Hao Chen, Yu~Chen, Wei Yin, and Chunhua Shen.
\newblock Generic perceptual loss for modeling structured output dependencies.
\newblock In \emph{Proceedings of the IEEE/CVF Conference on Computer Vision
  and Pattern Recognition}, pages 5424--5432, 2021.

\bibitem[Loshchilov and Hutter(2017)]{loshchilov2017sgdr}
Ilya Loshchilov and Frank Hutter.
\newblock Sgdr: Stochastic gradient descent with warm restarts, 2017.

\bibitem[Loshchilov and Hutter(2019)]{loshchilov2019decoupled}
Ilya Loshchilov and Frank Hutter.
\newblock Decoupled weight decay regularization, 2019.

\bibitem[Menze et~al.(2014)Menze, Jakab, Bauer, Kalpathy-Cramer, Farahani,
  Kirby, Burren, Porz, Slotboom, Wiest, et~al.]{menze2014multimodal}
Bjoern~H Menze, Andras Jakab, Stefan Bauer, Jayashree Kalpathy-Cramer, Keyvan
  Farahani, Justin Kirby, Yuliya Burren, Nicole Porz, Johannes Slotboom, Roland
  Wiest, et~al.
\newblock The multimodal brain tumor image segmentation benchmark ({BRATS}).
\newblock \emph{IEEE Transactions on Medical Imaging}, 34\penalty0
  (10):\penalty0 1993--2024, 2014.

\bibitem[Oord et~al.(2018)Oord, Li, and Vinyals]{oord2018representation}
Aaron van~den Oord, Yazhe Li, and Oriol Vinyals.
\newblock Representation learning with contrastive predictive coding.
\newblock \emph{arXiv preprint arXiv:1807.03748}, 2018.

\bibitem[Russakovsky et~al.(2015)Russakovsky, Deng, Su, Krause, Satheesh, Ma,
  Huang, Karpathy, Khosla, Bernstein, et~al.]{russakovsky2015imagenet}
Olga Russakovsky, Jia Deng, Hao Su, Jonathan Krause, Sanjeev Satheesh, Sean Ma,
  Zhiheng Huang, Andrej Karpathy, Aditya Khosla, Michael Bernstein, et~al.
\newblock Imagenet large scale visual recognition challenge.
\newblock \emph{International journal of computer vision}, 115\penalty0
  (3):\penalty0 211--252, 2015.

\bibitem[Shen et~al.(2017)Shen, Wang, Zhang, and McKenna]{shen2017boundary}
Haocheng Shen, Ruixuan Wang, Jianguo Zhang, and Stephen~J McKenna.
\newblock Boundary-aware fully convolutional network for brain tumor
  segmentation.
\newblock In \emph{International Conference on Medical Image Computing and
  Computer-Assisted Intervention}, pages 433--441. Springer, 2017.

\bibitem[Simonyan and Zisserman(2015)]{simonyan2015deep}
Karen Simonyan and Andrew Zisserman.
\newblock Very deep convolutional networks for large-scale image recognition,
  2015.

\bibitem[van~der Voort et~al.(2021)van~der Voort, Incekara, Wijnenga, Kapsas,
  Gahrmann, Schouten, Dubbink, Vincent, van~den Bent, French,
  et~al.]{van2021erasmus}
Sebastian~R van~der Voort, Fatih Incekara, Maarten~MJ Wijnenga, Georgios
  Kapsas, Renske Gahrmann, Joost~W Schouten, Hendrikus~J Dubbink, Arnaud~JPE
  Vincent, Martin~J van~den Bent, Pim~J French, et~al.
\newblock The erasmus glioma database (egd): Structural mri scans, who 2016
  subtypes, and segmentations of 774 patients with glioma.
\newblock \emph{Data in brief}, 37:\penalty0 107191, 2021.

\bibitem[Zeng et~al.(2021)Zeng, Wu, Hu, Xu, Yuan, Huang, Zhuang, Hu, and
  Shi]{zeng2021positional}
Dewen Zeng, Yawen Wu, Xinrong Hu, Xiaowei Xu, Haiyun Yuan, Meiping Huang, Jian
  Zhuang, Jingtong Hu, and Yiyu Shi.
\newblock Positional contrastive learning for volumetric medical image
  segmentation.
\newblock In \emph{International Conference on Medical Image Computing and
  Computer-Assisted Intervention}, pages 221--230. Springer, 2021.

\bibitem[Zhao et~al.(2017)Zhao, Shi, Qi, Wang, and Jia]{zhao2017pyramid}
Hengshuang Zhao, Jianping Shi, Xiaojuan Qi, Xiaogang Wang, and Jiaya Jia.
\newblock Pyramid scene parsing network.
\newblock In \emph{Proceedings of the IEEE conference on computer vision and
  pattern recognition}, pages 2881--2890, 2017.

\bibitem[Zhou et~al.(2019)Zhou, Sodha, Siddiquee, Feng, Tajbakhsh, Gotway, and
  Liang]{zhou2019models}
Zongwei Zhou, Vatsal Sodha, Md~Mahfuzur~Rahman Siddiquee, Ruibin Feng, Nima
  Tajbakhsh, Michael~B Gotway, and Jianming Liang.
\newblock Models genesis: Generic autodidactic models for 3d medical image
  analysis.
\newblock In \emph{Medical Image Computing and Computer-Assisted Intervention},
  pages 384--393, 2019.

\end{thebibliography}

\newpage
\appendix
\section{}
\subsection{Reconstruction results}

\begin{figure}[!ht]
    \centering
    \includegraphics[width=0.95\textwidth]{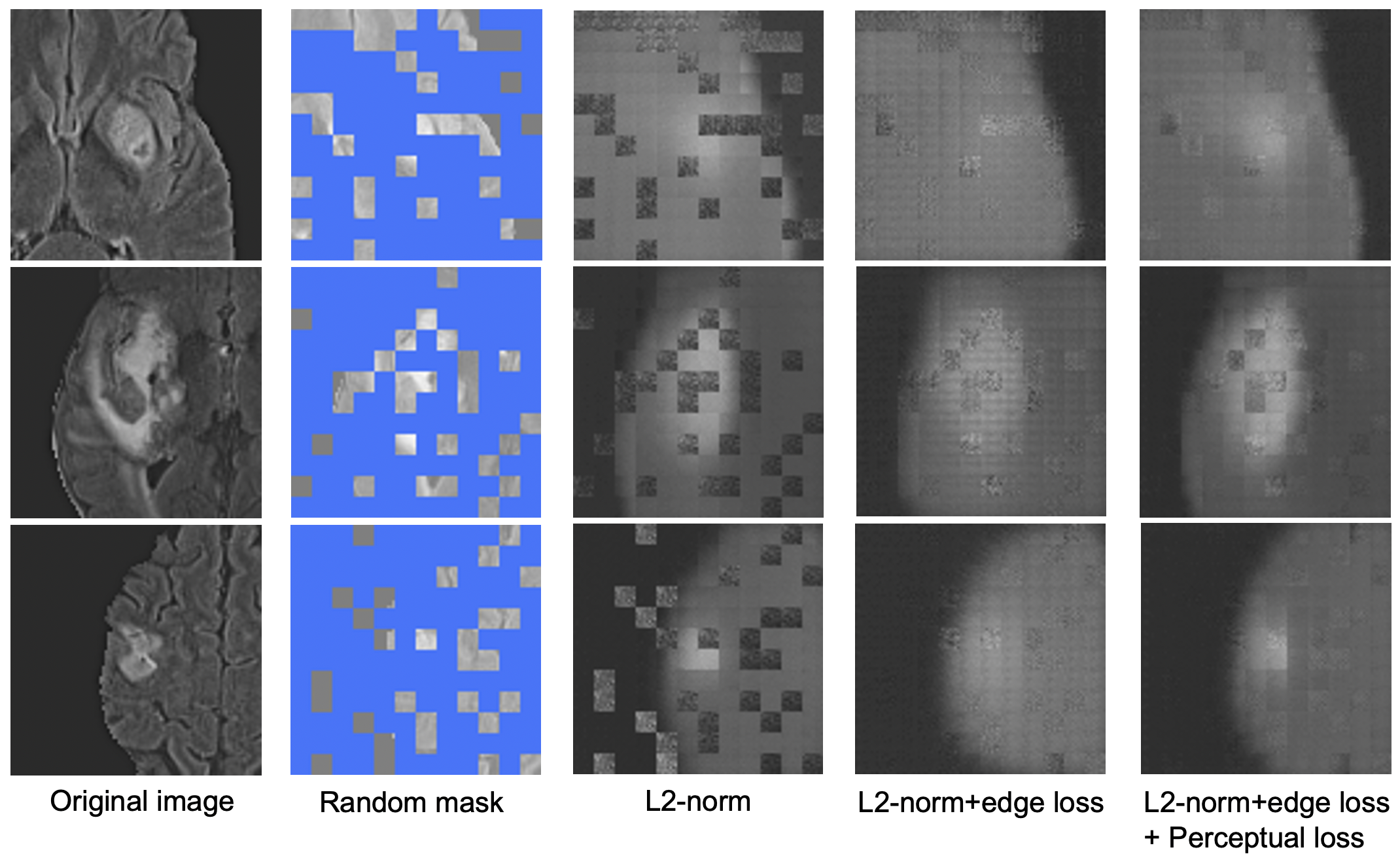}
    \caption{\small Effect of different loss functions for image reconstruction. We observe the progressive improvement of the reconstructed results from a fraction (25\%) of the input.}
    \label{fig:reconstruction}
\end{figure} 

\begin{table*}[!ht]
  \caption[table: Comparison_all_methods]{\small Results of two qualitative metrics on image quality of reconstruction results generated by different methods in Fig.~\ref{tab:results_qualitative}. The values are computed between the reference image and the reconstructed image.}
  \label{tab:results_qualitative}
  \centering
  \setlength{\tabcolsep}{2mm}{
   \begin{tabular}{l | c | c | c }
    \toprule

    &\multicolumn{1}{c|}{\textit{Sample 1}} & \multicolumn{1}{c|}{\textit{Sample 2}} & \multicolumn{1}{c}{\textit{Sample 3}}\\
    {Methods}
 
     ~ &  \textit{PSNR/SSIM} &  \textit{PSNR/SSIM}  &  \textit{PSNR/SSIM}     \\
     \hline
     $\mathcal{L}_{2}$ norm & 21.475/0.396 & 24.064/0.558  & 25.362/0.521 \\
     $\mathcal{L}_{2}$+$\mathcal{L}_{edge}$   & 24.497/0.503 & 26.751/0.565 &26.891/0.571\\
     $\mathcal{L}_{2}$+$\mathcal{L}_{edge}$+$\mathcal{L}_{per}$  & \textbf{25.339/0.514} & \textbf{27.211/0.572} & \textbf{27.703/0.586}\\

\bottomrule
  \end{tabular}
}
\end{table*}

\subsection{3D Positional Encoding}
\begin{figure}[!ht]
	\begin{center}
		\includegraphics[width=0.8\textwidth]{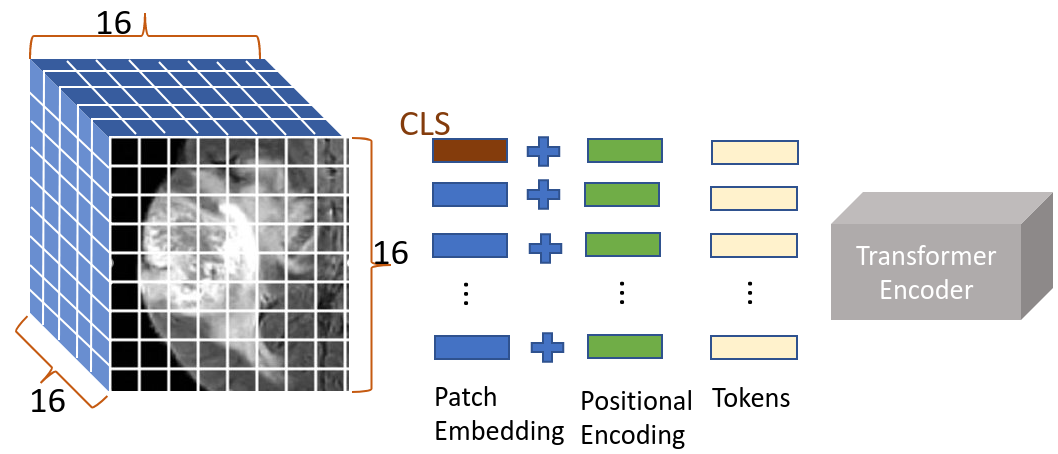}
	\end{center}
    	\caption{Position encoding representation. The given input volume is divided into a grid of size 16 $\times$ 16 $\times$ 16. A special CLS token is added along with the patch embeddings. 3D sinusoidal position encoding is added to each input patch (including the special CLS token). The position-aware tokens are fed into the Transformer encoder.}
	\label{fig:pos_encod} 
\end{figure}

In our setup, ViT deals with 96 $\times$ 96 $\times$ 96 cubes. The input volume is divided into patches of 8 $\times$ 8 $\times$ 8. A special \textit{CLS} token is added to the input sequence (see Fig. \ref{fig:pos_encod}). This token is used for downstream calculations. Fixed 3D sinusoidal encoding is used to inject positioning information to these patches, including the \textit{CLS} token. There are 1729 input tokens (1728 patch tokens and one \textit{CLS} token). Each axis encodes approximately $\frac{1}{3}$ of the input volume. The framework is flexible to work with different patch sizes.

\subsection{Perceptual loss and edge loss}
\begin{figure}[!ht]
	\begin{center}
	\includegraphics[width=0.70\textwidth]{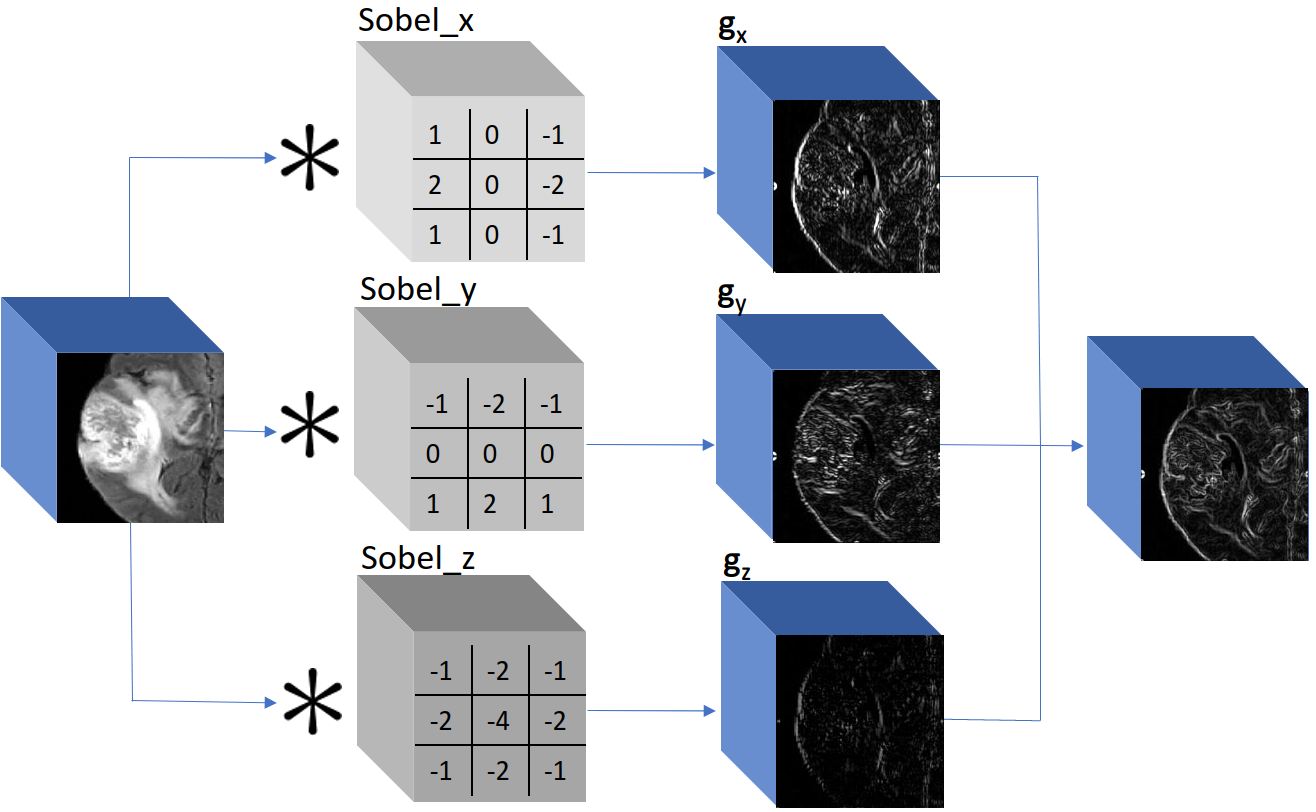}
	\end{center}
    	\caption{Schematic diagram of 3D Sobel filter. The 3D volume is convolved by 3 $\times$ 3 filters. The filter weights are fixed, and each filter computes the gradients in one specific direction. Sobel\_x computes the gradients $g_x$ along the Sagittal plane (x-axis). Similarly, $g_y$ and $g_z$ are gradients along the Axial (y-axis) and Coronal (z-axis) planes, respectively. The final edge map is obtained by $\sqrt{g_x^2 + g_y^2 + g_z^2}$.}
	\label{fig:sobel} 
\end{figure}

For medical data, a pre-trained VGG-16 \cite{simonyan2015deep} is used to compute the perceptual similarity between the original volume and its reconstruction. The VGG network is pre-trained on 2D images and can not be directly used with 3D volumes. Hence, we compute loss along 2D slices of the input volume and its reconstruction. The per-slice loss values are averaged to obtain loss for the entire volume. The 3D $Sobel$ filter operation is shown in Fig. \ref{fig:sobel}.

\subsection{Sample X-ray images.}

\begin{figure}[!ht]
    \centering
    \includegraphics[width=0.95\textwidth]{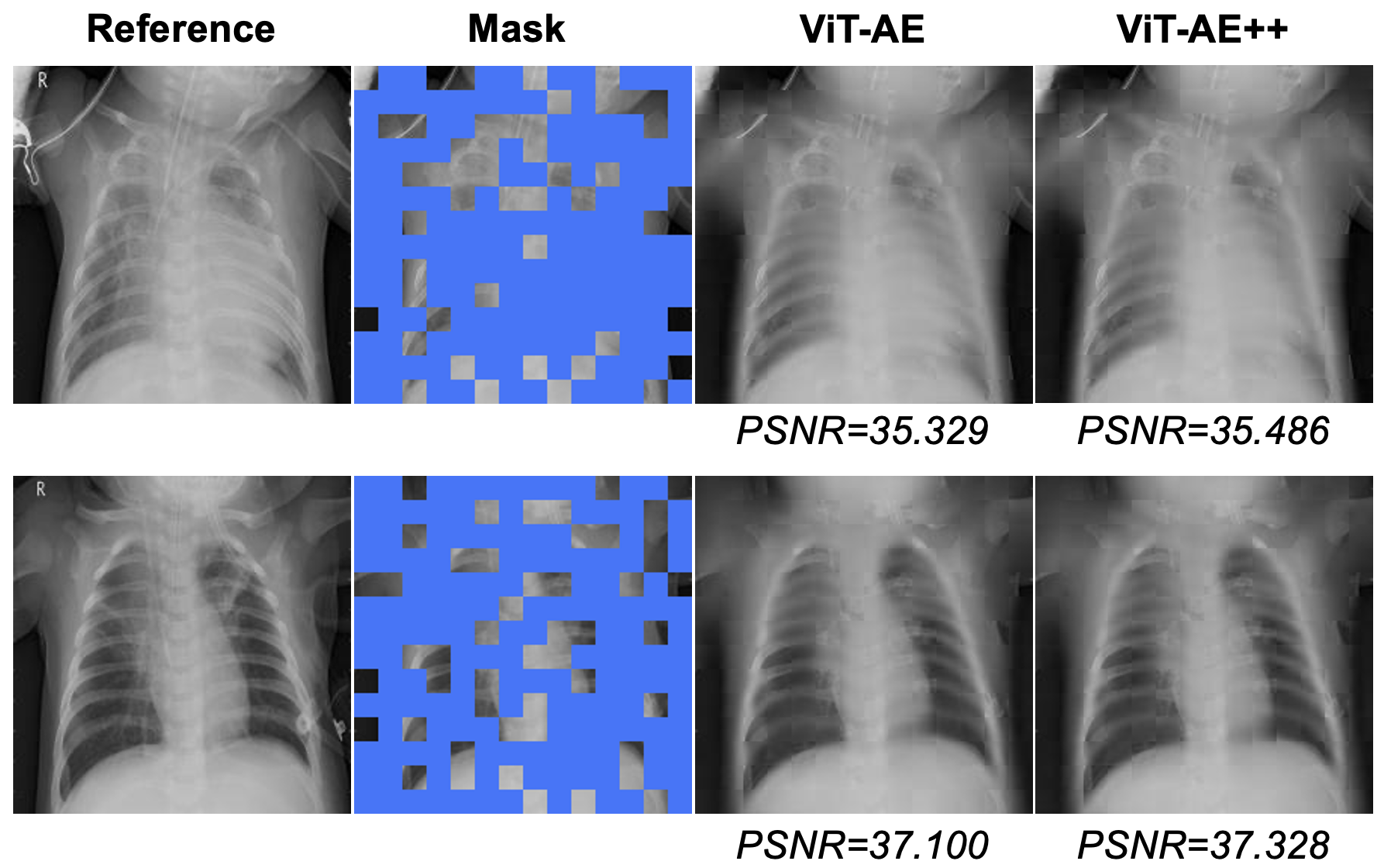}
    \caption{\small We observed that ViT-AE++ achieved significantly higher PSNR (p-value < 0.0001) than ViT-AE. There is no statistical significance in SSIM. }
    \label{fig:reconstruction_psnr}
\end{figure} 

\end{document}